\documentclass[journal]{IEEEtran}
\ifCLASSINFOpdf
\else
\fi
\usepackage{cite}
\usepackage{amsmath,amssymb,amsfonts}
\usepackage{graphicx}
\usepackage{textcomp}

\usepackage{booktabs}
\usepackage{amsmath}
\usepackage{amssymb}
\usepackage{amsfonts}
\usepackage{graphicx}
\usepackage{amsthm}
\usepackage{xcolor}
\usepackage{mathtools}
\usepackage{float}
\usepackage{pgfplots}
\pgfplotsset{compat=1.7}
\usepackage[vlined, ruled, shortend, linesnumbered]{algorithm2e}

\usepackage[style=base]{caption}
\usepackage{subcaption}

\usepackage{hyperref}
\usepackage{tabstackengine}
\setstackEOL{\cr}
\usepackage{flushend}

\usepackage{comment} 
\usepackage{todonotes} 
\usepackage{ifthen} 
\usepackage{gensymb}
\usepackage{siunitx}
\usepackage{placeins}

\usepackage[style=base]{caption}

\begin{document}
%
\title{Localization in Unstructured Environments:  Towards Autonomous Robots in Forests \\ with Delaunay Triangulation}
%
%
%

\author{Qingqing Li,
        Paavo Nevalainen,
        Jorge Pe\~{n}a Queralta,
        Jukka Heikkonen
        and Tomi Westerlund
    \thanks{Qingqing Li, Paavo Nevalainen, Jorge Pe\~{n}a Queralta, Jukka Heikkonen
        and Tomi Westerlund are with the Turku Intelligent Embedded and Robotic Systems (TIERS) Lab, University of Turku, Turku, Finland, e-mails: \{qingqli, ptneva, jopequ, jukhei, tovewe\}@utu.fi}
}

\maketitle


\begin{abstract}
    Autonomous harvesting and transportation is a long-term goal of the forest industry. One of the main challenges is the accurate localization of both vehicles and trees in a forest. Forests are unstructured environments where it is difficult to find a group of significant landmarks for current fast feature-based place recognition algorithms. This paper proposes a novel approach where local observations are matched to a general tree map using the Delaunay triangularization as the representation format. Instead of point cloud based matching methods, we utilize a topology-based method. First, tree trunk positions are registered at a prior run done by a forest harvester. Second, the resulting map is Delaunay triangularized. Third, a local submap of the autonomous robot is registered, triangularized and matched using triangular similarity maximization to estimate the position of the robot. We test our method on a dataset accumulated from a forestry site at Lieksa, Finland. A total length of 2100\,m of harvester path was recorded by an industrial harvester with a 3D laser scanner and a geolocation unit fixed to the frame. Our experiments show a 12\,cm s.t.d. in the location accuracy and with real-time data processing for speeds not exceeding 0.5\,m/s. The accuracy and speed limit is realistic during forest operations. 
\end{abstract}


%
\IEEEpeerreviewmaketitle

\section{Introduction}

Over the last ten years, autonomous harvesting and transportation have become the long-term goal of the future development of the forest industry and attracted the interest of the research community. Naturally, there are several intermediate goals such as the organization of the public data~\cite{kankare2015}, and the gradual increase of autonomy in varying degrees, starting with short-range transport in forests~\cite{hellstrom2009}. Incremental advances in forest autonomy include driver assistance platforms and function-specific automation. For example, these include the automated selection of tree stems to be processed, micro-tasks such as sequencing the processing of individual trees, local route planning, or semi-autonomous transportation and quality assurance. In all these tasks, one key element is the availability of tree maps, together with methods enabling the identification and selection of individual trees. Besides, local route planning requires accurate updates on the position of the harvester inside the forest in real-time, which can not be relied on global navigation satellite system (GNSS) sensors.

This work focuses on the real-time localization of an autonomous forwarder unit, a forestry vehicle that collects felled logs and hauls them to a local loading area. The required transport distance is usually short, in the range of 100-400\,m, and happens along a rough-terrain track a forest harvester has previously made while performing cut-to-length logging. To autonomously navigate in the forest, visual sensors present significant challenges owing to the lack of a stable background from which contours can be extracted, as well as demanding low-light and harsh weather conditions~\cite{yoneda2019automated}. For this type of environment, with mostly empty space, clearly separated and mostly homogeneous trees, ranging sensors show a wider potential. In particular, in our work, we rely on Light Detection And Ranging (lidar), a viable technology for this environment, with commercial sensors able of centimeter-level accuracy, over 100\,m range, and wide field-of-view.

There are multiple well-established frameworks and algorithms for autonomous driving in urban environments~\cite{badue2019self}, as well as localization and mapping in roads and buildings~\cite{thakur2016scanning}. Many of the solutions proposed in these areas have a strong dependency on lidar scanners~\cite{qingqing2019multi}, among other sensors. In the field of forest mapping and navigation, several researchers have utilized terrain laser scan (TLS) to build point-cloud maps~\cite{zhang2019novel, pierzchala2018, liang2016, miettinen2007, tang2015slam}. In some of these works~\cite{zhang2019novel, liang2016}, data is collected from fixed-positioned tripods in order to gain an understanding of how well individual trees can be detected and their diameter at the breast height (DBH) can be registered. Other studies rely on mobile laser scanning (MLS)~\cite{pierzchala2018, miettinen2007, tang2015} to simulate the operation of the harvester. 

Navigating in a forest presents several inherent challenges owing to the complexity and lack of structure in the environment. A realistic forest harvesting process involves constant obstructions from cut or fell branches and trunks and irregular terrain. Furthermore, the movement of a harvester tends to be rotational most of the time, with constant changes in orientation along the work cycle and very slow translational motion or even with no movement at all while trees are being cut. These particularities of forest navigation bring both advantages and disadvantages to standard development of autonomous mobile robots: idle periods and slow motion aid at realizing real-time operation and data processing, while fast rotations difficult accurate mapping. A small error in the orientation estimation can significantly affect the mapping of trees that are farther away.

Research in the field of localization and mapping for autonomous robots can be roughly divided into two approaches: simultaneous localization and mapping (SLAM), and sequential location and mapping. The main difference between these two is that SLAM has the location and mapping occurring at the same time, whereas the sequential version has the mapping part completed after the track of the vehicle has been constructed for a considerable length. Localization is often done by finding either a partial match between sensor data batches acquired at different times, or finding a spatial transformation that maximizes the data matching. A different approach is then to classify localization methods based on whether the matching process is done globally or locally. The same can be applied to different types of sensor data, including images~\cite{sattler2016efficient}. Nonetheless, we are interested in matching point clouds generated from lidar scanners. Because locally converging methods have a valley of convergence~\cite{magnusson2009}, which is sometimes defined e.g. by step length limit and the point cloud overlap limit, we study a global point cloud matching approach.

In this paper, our objective is to enable fast, real-time global localization for autonomous forest harvesters and forwarders. Our experiments have been done using data from a 3D lidar scanner situated in the front panel of a manually operated forwarder. The methods presented in this paper rely on the assumption that a global forest map in the form of a point cloud is available. In practice, we first create the map using state-of-the-art lidar odometry and mapping algorithms, and then evaluate our localization approach. The localization is global and does not depend on previous states or position estimations, therefore providing a robust solution for long-term autonomy.

\subsection{Localization in Autonomous Mobile Robots}
Autonomous mobile robots meant to operate outdoors often rely on GNSS sensors as the basic source of global localization data, and then integrate other sensor data through sensor fusion techniques for local position estimation~\cite{qingqing2019multi}. GNSS sensors alone do not provide enough accuracy in urban or dense environments, with accuracy often in the order of meters~\cite{tomavstik2016horizontal}. While GNSS sensors have increased their accuracy in recent years, multiple challenges remain in environments with structures affecting the signal path. In particular, GNSS signals are weak in forests, owing to the irregular land contours and the coverage that tree foliage provides~\cite{zimbelman2018}. Therefore, we focus on the design and development of localization methods that rely solely on lidar data, with GNSS providing only the initial position before entering the forest or starting the autonomous operation. The following are the main approaches enabling the estimation of the movement of a mobile robot based on lidar scans by either comparing consecutive scans (hereinafter also called point clouds) or a given scan with a global point cloud. We refer to comparing scans as scan matching:

\textbf{Full scan matching.} In SLAM algorithms, a necessary step previous to update the map is to estimate the relative movement of the robot with respect to its last known position. This is equivalent to finding a geometrical transformation between the corresponding point clouds. One of the most widely utilized methods for calculating such transformation is the iterative closes point (ICP) algorithm~\cite{rusinkiewicz2001efficient, segal2009generalized}. In ICP, pairs of points between the two point clouds being compared are iteratively selected until a transformation with acceptable pairing error distance is found. Several variants of the ICP algorithm are available through the open-source point cloud library (PLC)~\cite{holz2015registration}. Other popular methods include the perfect match (PM)~\cite{lauer2005perfectmatch}, or normal distribution transforms (NDT)~\cite{biber2003ndt} algorithms. Most of these algorithms can be extended and utilized for map-based localization, where a global point cloud is available and a given lidar scan is matched with only a subset of the global map. However, in these cases, an estimation of the previous position is needed, since global point cloud matching might be unfeasible due to the computational complexity. Moreover, ensuring convergence in global matching presents significant challenges. Since our objective is to enable global localization without dependence on previous states, we develop a novel method that takes into account the geometry of the environment instead of using the full point cloud.

\textbf{Feature-based matching.} Generic point cloud matching algorithms do not take into account the structure of any potential features within the scans that are being compared. When the environment where robots operate has known structures, feature-based scan matching can significantly enhance the accuracy of the matching process~\cite{nunez2006feature, sampath2006clustering, liang2011new}. Moreover, owing to the preprocessing step in which raw data is transformed into features, the computational time required to calculate the transformation can be reduced. In this direction, Zhang \textit{et al.} presented the lidar odometry and mapping (LOAM) method~\cite{zhang2014loam}, which assumes a structured environment where planes and edges can be detected. Then, the transformation between two point clouds can be estimated by aligning the planar and edge feature points from each of them. Multiple algorithms have extended the original LOAM method for other types of features. Among them, Shan \textit{et al.} presented the ground-optimized Lego-Loam~\cite{shan2018lego}, which delivers optimized real-time three-dimensional point cloud matching in small scale embedded devices. While feature-based matching algorithms yield accurate results, their applicability is limited in forests, with most methods focusing on urban or indoor environments. In forests or other unstructured environments, edge and planar features are either noisy or missing. In this paper, we still rely on the state-of-the-art Lego-Loam algorithm for building a map of the forest prior to the operation of the robots. This process is done offline. However, we develop an alternative method for localization because of our needs for real-time operation and global localization.

\textbf{Geometrical matching methods.} Geometry-based localization algorithms have the potential to recognize the position in constant time, which is an attractive property towards real-time localization with embedded processors. Geometrical methods are often applied to graph-like structures to find geometric transformations that match either the complete graph or a set of subgraphs. Thrun \textit{et al.} developed the notions of sparse extended information filters (SEIFs) which exploit the structure inherent through the local web-like networks of features maps~\cite{thrun2002slam, liu2003results}. In a similar direction, Ulrich \textit{et al.} proposed a global topological representation of places with object graphs serving as local maps for place classification and place recognition~\cite{ulrich2000appearance, chen2010enhanced, himstedt2014large}. Among the most relevant approaches to our work are place-recognition algorithms, such as Bosse's work on a keypoints similarities voting method in 2D and 3D lidar data~\cite{lynen2017trajectory, bosse2013place, bosse2009keypoint}. 


\subsection{Contribution and Structure}

Most of the aforementioned point cloud matching algorithms focus on structured urban environments, or trajectory tracking through fixed, predefined paths such as roads. Nonetheless, their performance is significantly degraded in unstructured environments and, in particular, in forests. In a forest harvesting operation, harvesters and forwarders travel through undefined paths. Moreover, the detection of potential features (ground and tree stems) is considerately affected by irregularities (branches and foliage in the trees, and small plants, rocks or fell trees in the ground). While some of the state-of-the-art methods yield reasonably good results for mapping and odometry, our aim is to develop a fast method for global localization able to run in real-time and taking into account the known geometrical properties of forest environments. Therefore, in this work, we pursue the design and development of a localization method which is reliable, accurate and of low computational cost. The computation is aimed to occur in real time onboard of harvesters and forwarders in the forest. Relatively randomly located but ubiquitous tree stems are a natural basis for localization efforts. We also assume that a map of the forest in the form of a point cloud is available before the mission starts. 


We propose a lightweight and geometry-based localization method for pose estimation within a global map of a harvester in dense and unstructured forest environments. The proposed method is lightweight and matches a triangularization of a single lidar scan with a subgraph of the triangularization of the global forest map. Compared to matching raw lidar data point-by-point, we reduce the problem to a triangularization created based on the positions of tree stems. This allows us to reduce the amount of point to be matched to approximately 1\% of the size of a single lidar scan because individual trees get an average of 100 laser hits in our setting: a 16-channel 3D lidar with a field of view of about \ang{210} and situated at a height of 1.5\,m. Closer trees get, however, considerably higher amounts of laser hits. The proposed method is based on triangle dissimilarity minimization, and that is why we say the method is geometry-based. To the best of our knowledge, this is the first paper to utilize Delaunay triangulation for the purpose of localizing a vehicle in a forest environment. Moreover, we provide a fully experimental approach with a realistic use case in forest harvesting. The main contributions of our work are the following:
\begin{itemize}
  \item the introduction of a Delaunay Triangulation graph map for localization in forests; and
  \item the design and development of a framework to solve the vehicle tracking problem in a local coordinate system relying solely on lidar data, without GNSS or inertial sensor data.
\end{itemize} 

The remainder of this paper is organized as follows. Section 2 introduces the study area and equipment, the optimization problem to be solved, and the methodology we have followed. Section 3 reports experimental results. In Section 4, we discuss the potential improvements, as well as the main benefits of our approach. Finally, Section 5 concludes the work and outlines future research directions.

\begin{figure*} 
    \centering
    \includegraphics[width=0.9\textwidth]{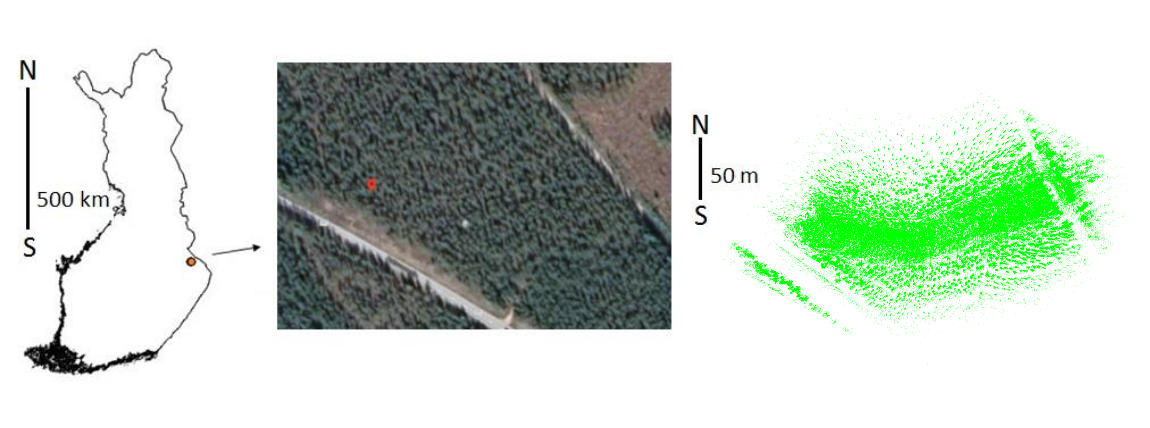}
    \caption{ (\textbf{a})~The test site in Lieksa, Eastern Finland. 
(\textbf{b})~The forest canopy map from Google maps. (\textbf{c})~Point cloud map of the study area.}
    \label{fig:study_area}
\end{figure*} 

\section{Material and Results}

\begin{figure} 
    \centering
    \includegraphics[width=0.45\textwidth]{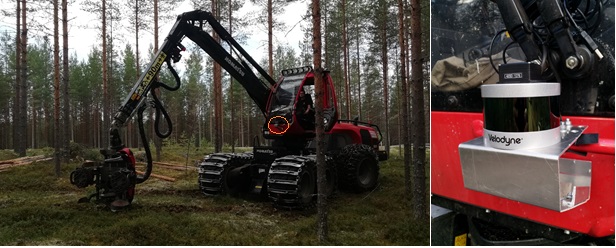}
    \caption{(\textbf{a}) Komatsu Forest 931.1 forest harvester. 
    The lidar scanner is marked by a red circle in the front of the 
    windshield of the cabin. 
  (\textbf{b}) A close-up of the lidar scanner.}
    \label{fig:lidarplacement}
\end{figure} 

\begin{figure}
     \centering
    \includegraphics[width=0.45\textwidth]{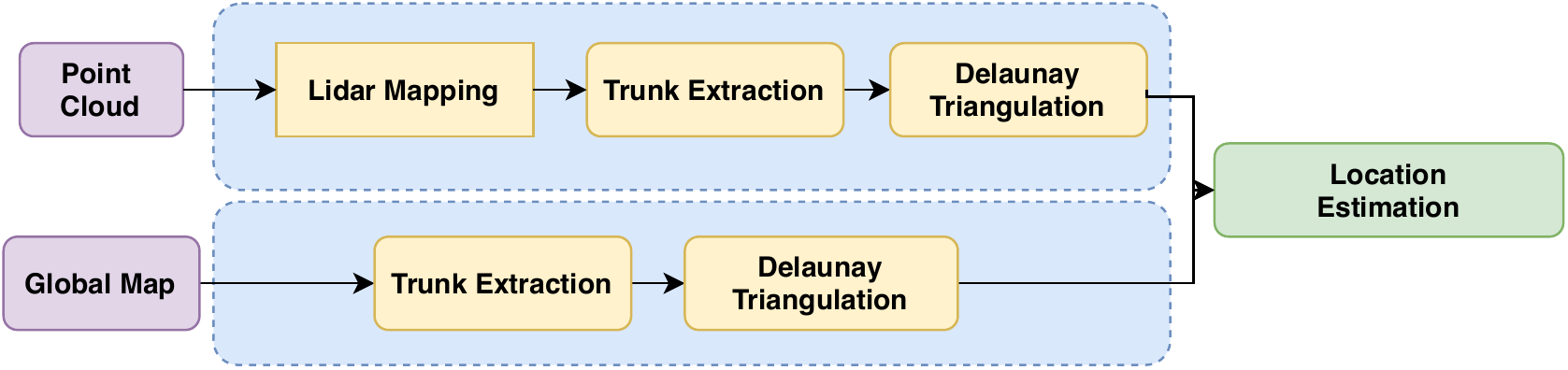}
    \caption{System overview: sensor data is matched with a global map to produce a global position estimation. Consecutive runs are independent.}
    \label{fig:sys_overview}
\end{figure}
 
 \begin{figure*}
    \centering
    \includegraphics[width=0.8\textwidth]{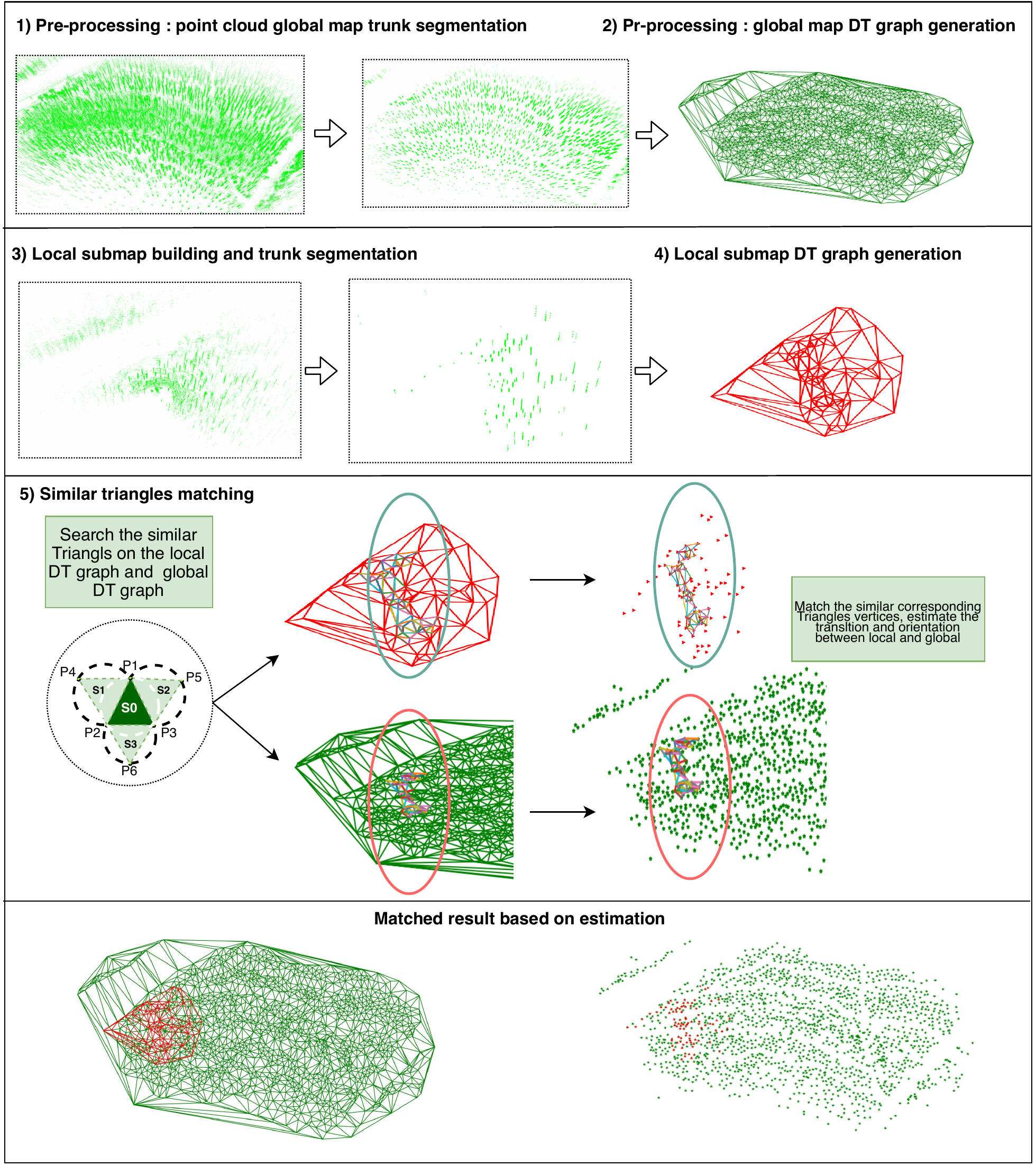}
    \caption{Schematic representation of the proposed method to predict the transformation between local observation submap and global map.}
    \label{fig:schematic_overview}
\end{figure*} 

\subsection{Study area}

The study area covers one mature pine stand ready for second thinning, which represents a typical second thinning forest for Finnish standards. The location of the study area is illustrated in Figure~\ref{fig:study_area}. The terrain profile was mostly flat, the maximum height difference over the site was 8 m. The overall trail length is approximately 1200\,m covering roughly 2200\,trees. 

\subsection{System Hardware}

The mobile platform was a Komatsu Forest 931.1 forest harvester with a GPS unit and the lidar unit attached to the top of the cabin. The harvester has physical dimensions (length, width, height) of 7.6\,m by 2.9\,m by 3.9\,m, and its mass is 19610 kg.The maximum driving speed is 8\,km/h off-road and 25\,km/h on road.

The on-board lidar is Velodyne-16, a 16 channel lidar with 3cm distance accuracy, \ang{360} horizontal field of view, \ang{30} vertical field of view with $\pm 15^o$ up and down, 5-20\,Hz scanning frequency and 100\,m scan range. As Figure~\ref{fig:lidarplacement} shows, the lidar is fixed on the front of the harvester with a \ang{210} horizontal view. The harvester has a folding frame, which means the point cloud frames scanned have constantly alternating horizontal orientation during the work cycle.

The GPS data was recorded by Spectra SP60 GPS unit. This unit fully utilized all 6 GNSS systems: GPS, GLONASS, BeiDou, Galileo, QZSS and SBAS. In SBAS (WAAS/EGNOS/MSAS/GAGAN) mode, the horizontal position error smaller than 50\,cm, and the vertical error is smaller than 85\,cm. In differential GPS mode, the accuracy is able to reach 25\,cm in horizontal and 50\,cm in vertical accuracy. In Real-Time Kinematic position (RTK) mode, the accuracy can reach 8\,mm in horizontal and 15\,mm in vertical accuracy. Nonetheless, these values are not achievable under dense foliage in a forest environment. Regarding the computing platform, the proposed methods have been tested on an Intel Core i7-9700K CPU (8 cores, up to 3.60GHz), and 16\,GB of RAM.  

\subsection{Methodology} 

Figure~\ref{fig:sys_overview} shows an overview of our proposed method. The complete system takes as a sole input the point cloud data from the 3D lidar, and outputs the position and orientation estimation in the reference of the given global map. The system can be divided into six steps as illustrated in Figure~\ref{fig:schematic_overview}. The former two steps are done offline by processing the point cloud data defining the forest map. The latter four steps are then done online onboard the harvester to estimate its position in real time. The steps are:

\begin{enumerate}
    \item Point-cloud trunk segmentation from the global map (offline).
    \item Delaunay triangulation of the global map from the segmented trunk points (offline).
    \item Aggregation of 3D lidar scans into a local point cloud defining the robot's position (online).
    \item Segmentation of trunks from the local point cloud (online).
    \item Delaunay triangulation from the local segmented trunk points (online).
    \item Estimation of a geometrical transformation that matches the local Delaunay triangulation with a subset 
\end{enumerate}

A 2D Delaunay triangulation (DT) process~\cite{edelsbrunner2000} takes in a set of planar points $V \subset \mathbb{R}^2$ and produces a triplet $G=(V,E,T)$, where $E \subset V^2$ is a set of edges and $T \subset V^3$ is a set of triangles with the so called Delaunay property i.e. the circumscribed triangle associated with each triangle $t \in T$ contains no points $v\in V$ others than the three vertex points of the triangle $t$. 

The first step in the proposed method takes a global point cloud $PC_{map}= \{P_i\}_{i=1..n_m} \subset \mathbb{R}^3$, representing the map of the forest, and produces a robust set of trunk positions $PC_{trunks} = \{ P_i\}_{i=1...n_t} \subset \mathbb{R}^2$. Note that $PC_{trunks}$ is not a subset of $PC_{map}$. The second step in our method then subjects the horizontal plane projection $PC_{map}$ to the Delaunay triangularization graph $G_{map} = (V_{map}, E_{map}, T_{map})$ for matching. As we mentioned earlier, both if this steps happen offline before the robot is deployed, or every time the robot gets a global map update.

The third, fourth and fifth steps, which happen online, cover the generation of a local, real-time Delaunay triangulation of the trees within the field of view of the robot. First, we accumulate and aggregate several raw point cloud scans from the 3D lidar and generate a local point cloud $PC_{local} \subset \mathbb{R}^3$. The aggregation relies on real-time lidar odometry from LOAM~\cite{zhang2014loam}. Then, following the same procedure as with the global map, we generate a robust two-dimensional set of trunk positions $L_{trunks} \subset \mathbb{R}^2$. From the set of trunks, we can define the local DT graph $G_{local}= (V_{local}, E_{local}, T_{local})$.

Finally, in the sixth step in the process we calculate a rigid body transformation, defined by a rotation $\theta$ and a translation $p_t$, between the local DT graph $G_{local}$ and a subset of the global DT graph $G_{match} \subset G_{map}$. The transformation relates directly to the robot position and orientation in the global map. Instead of matching large quantities of the 3D point cloud, the proposed system seeks to match a triangularized representation of the local data against a similar precomputed triangularized global representation. The resulting process is computationally efficient and with a low memory demand.

In the rest of this section, we further describe the process outlined above and delve into the details of the most critical steps. In general, the key idea is building a unique 2D graph topology from the 3D point cloud map, and finding the best match between local and global topology.

 \begin{figure*} 
    \centering
    \includegraphics[width=1.0\textwidth]{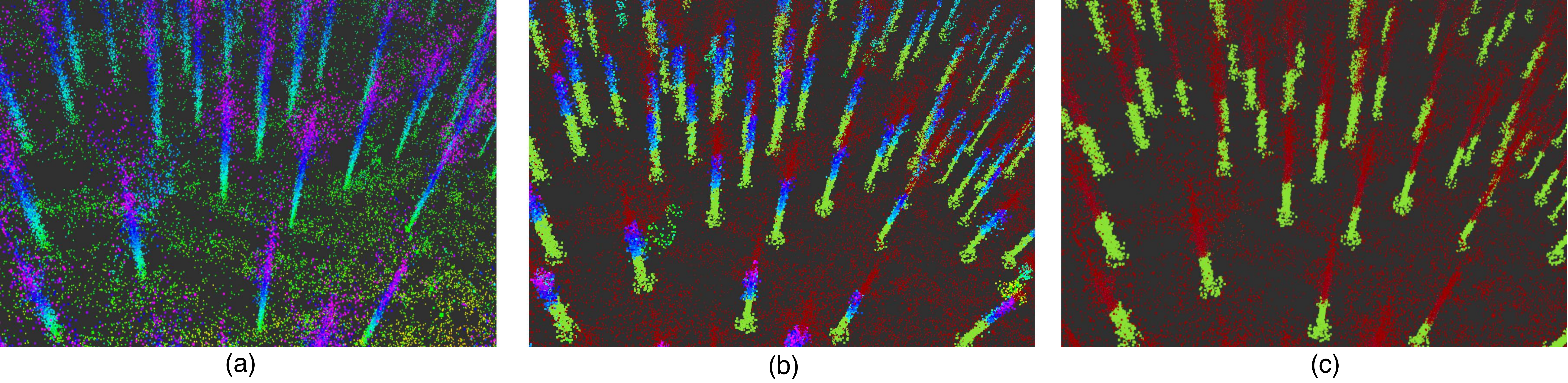}
    \caption{Trunk point cloud segmentation approach. 
    \textbf{(a)} The original point cloud map. 
    \textbf{(b)} The extracted trunk point cloud (blue and green)
    after removing ground and branch point cloud (red). 
    \textbf{(c)} The final extracted trunk point cloud (green) after clustering process} 
    \label{fig:trunks_seg}
\end{figure*} 

\subsection{Trunk Point Cloud Segmentation}\label{sec:map_trunkseg}

With a given map $PC_{map}$ of the forest, an essential step in the proposed method is to extract the trunks points that define $PC_{trunks}$ as a set of landmarks from $PC_{map}$. Compared to the other environmental features such as the bumpy terrain, branch structures, forest floor vegetation and the bushes in the forest, tree trunks are a significant and relatively noise-free feature.

To segment the trunk points, the first step is extracting the trunk point cloud $PC_{trunks}$ from the map point cloud $PC_{map}$ following the methodology described in Algorithm~\ref{alg:extractTrunkFromGmap}. We employ the Kd-Tree space partitioning structure to accelerate the neighborhood search~\cite{bentley1975multidimensional}, together with a Euclidean Cluster to find the trunk point cloud's~\cite{rusu2010semantic}. Instead of focusing on finding every trunk in the $PC_{map}$, we focus on extracting the most significant ones. We define the most significant trunks in this paper as those that show observational stability, i.e. they can be assumed to be easy to observe by the robot in the near future locations. Thus, we do not consider small trunks or bushes, which are not a worthwhile computational investment to locate and give them a landmark status. As a typical trunk is an approximately vertical object, we assume that the point from a stable trunk has a neighboring point located $2m$ above it. Then, we traverse all points in the $PC_{map}$, and for each point $p_i= (x_i,y_i, z_i)\in PC_{map}$ we try to find the nearest neighbor of a hypothetical point $p_i'=(x_i,y_i, z_i+t_h)$ which is $t_h$\,m above the point $p_i$. We add all such points $p_i$ to the set $PC_{trunks}$, which truly have near neighbors above them. Next, we form Euclidean clusters (each cluster connected by close distances between the members) and drop the clusters that have a size smaller than a threshold value $cluster\_threshold$. Finally, we compute the mean $\bar{p}\in\mathbb{R}^2$ of the horizontal projections of the cluster points to represent a landmark trunk. These landmarks then compose the 2D trunk map $M_{trunks}$. Figure \ref{fig:trunks_seg} show the proposed trunk point cloud segmentation result. The figure \ref{fig:trunks_seg}(a) shows the map to be processed, the figure \ref{fig:trunks_seg} (b) shows the extracted point cloud after removing the ground point cloud and branch point cloud from the input map. The figure \ref{fig:trunks_seg} (c) shows the final extracted stable trunk point cloud after cluster processing. 

\begin{algorithm}[t]
    \SetAlgoLined
    \KwData{A 3D point cloud map $PC_{map} = \{p_i\}_{i=1..n_m}$ }
    \KwResult{Trunk segmentation point cloud $PC_{trunks}= \{P_j\}_{j=1...N_t}$, \\ 
    2D Trunk map $M_{trunks} = \{ M_k \}_{k=1...n_t}$ }
    \ForEach{$p\in PC_{map}$}{ 
        $SearchPoint: p_{sp} = (p.x, p.y, p.z + h_{th})$\; 
        $NearestPointSearch: p_{sn} = nearestKSearch(P_m, p_{sp}) $ \;
        $ComputeDistance: distance(p_{sp}, p_{sn})$\;    
        \If{$ distance < d\_threshold $}{
            $P_t \leftarrow p_i$ // Add point to trunks point cloud\;
        } 
    }
    \ForEach{$cluster \in Clusters$}{
        \eIf{ $Cluster.points.size() < cluster\_threshold $}
        {Delete  $cluster$ from $Clusters$}
        {
            $\bar{p}= \mathbf{~mean~}_{p \in Cluster} p$ \;
            $M_{trunks} \leftarrow  \bar{p}$  \;
        }
    }
    \caption{Extracting trunks from the point cloud map}
    \label{alg:extractTrunkFromGmap}
\end{algorithm}

\subsection{Global Map DT Graph Generation}
\label{sec:Map_DT_triangulation}

A DT has multiple beneficial properties for our problem setting~\cite{lee1980two}, including the fact that it maximizes the minimal angle at all the vertices of the triangulation. This means that the noise in angular values is minimal in a noisy point set. A DT also defines the so called natural neighborhood around a point (the point set connected to a given point along the triangle edges), which solves the problem of setting the local point cloud feature scope or number of neighbors in an intuitive way~\cite{edelsbrunner2000}.

Our objective is to match the local trunk graph $G_{local}$ against a subset of the global trunk graph map $G_{map}$, and the prerequisite is that there exist some similar structures like triangles or small sets of triangles in both graphs. From the trunk landmarks $M_{trunks}$ of the $PC_{map}$ obtained as described in Section~\ref{sec:map_trunkseg}, we get the Delaunay triangulation $M_{trunks} \rightarrow G_{map}= (V_{map}, E_{map},T_{map})$ .

\subsection{Local Map and Point Cloud Aggregation}
    
In a dense forest environment, nearby trunks may be blocked from the lidar view. This may result in not enough stable tree trunks to match against the global map. Thus, in our case, we will employ the LOAM method, which builds a local map from aggregating several consecutive observations. However, we will also explore the direct construction using only one frame observation.

Both the construction of the ground truth map and the aggregation of consecutive lidar frames rely on the LOAM method~\cite{zhang2014loam}. It operates as follows. Let $P_t = \{p_i\}_{i=1..n}$ represent a raw scanned point cloud received at time $t$. All these raw point cloud's are processed with the LOAM algorithm to build a local map based on one scan or the aggregation of several consecutive scans. The LOAM algorithm is a state-of-the-art feature-based lidar odometry algorithm. LOAM receives the 3D point cloud from the lidar, and projects the point cloud onto a range image for feature extraction. By calculating the curvature and some features from each row of the range image, the registration process selects subsets $P_e$ and $P_p$ (edge and plane points, respectively). Instead of comparing all the points, LOAM utilizes only those two subsets to find a transformation between scans, and then a two-step Levenberg-Marquardt optimization method is employed to optimize the six-degree-of-freedom transformation across consecutive scans. The complete lidar odometry problem gets solved with a speed of 1 Hz resulting the local map $PC_{local}$ with the computing platform utilized in our experiments. 
    
In this study, the local point cloud $PC_{local}$ was generated with one or several scans by utilizing the estimated position from the LOAM method. The raw lidar output frequency is 30 HZ, but not every scan from the lidar stream needs to be used to calculate the robot odometry. To reduce the accumulation error and balance the computation burden, LOAM only considers those scans that the Euclidean distance between two observation positions is longer than a certain threshold distance (e.g. 30\,cm) or the angular change is larger than a certain threshold angle (e.g. \ang{30}). As the nearby trunks or objects may block a sector of the view, aggregating more frames observation from consecutive positions helps to increase the number of the stable trunks registered into the map $PC_{local}$.
    
\subsection{Local DT Graph Generation}
\label{sec:local_DT}

After we have obtained the local map $PC_{local}$, the next step is to generate the DT graph $G_{local} = \{V_{local},E_{local},T_{local}\}$, just as with the global map. The methodology explained in Section~\ref{sec:map_trunkseg} applies in this case as well, but this time producing a local map $L_{local}$, its 2D projected subset $L_{trunks}$, and a local DT graph $G_{local}$.
    
\begin{figure} 
    \centering
    \includegraphics[width=0.5\textwidth]{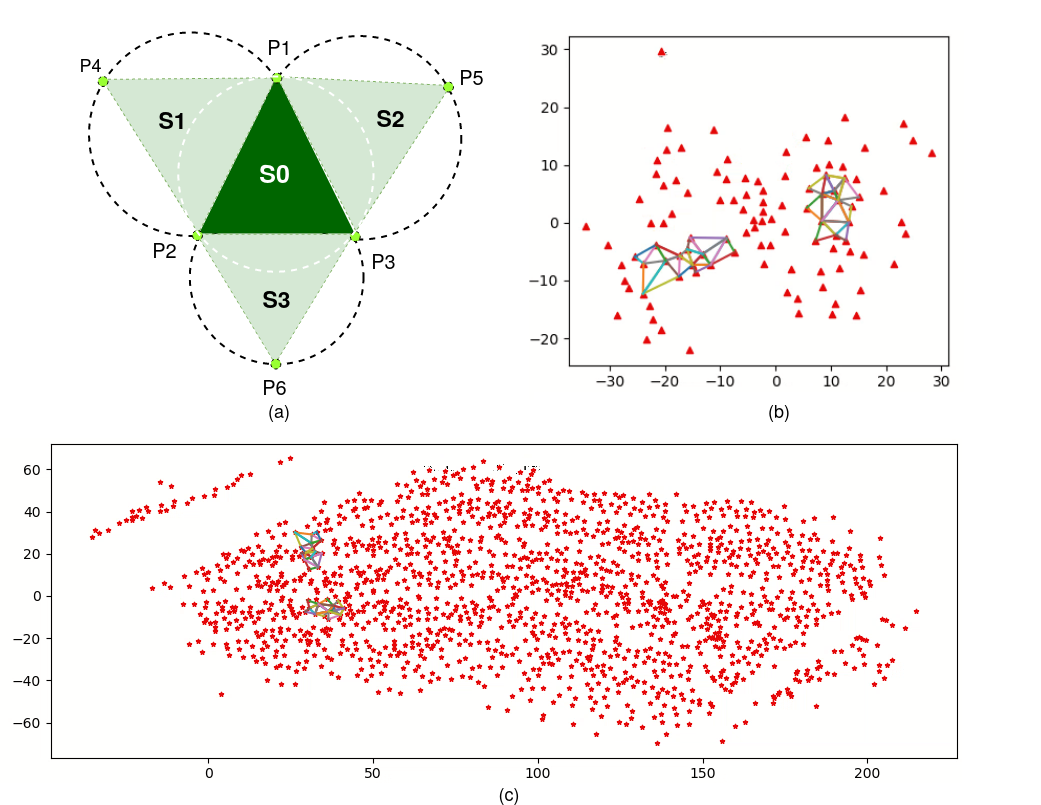}
    \caption{\textbf{(a)}: The target triangle $S_0$ with its
    neighbors $S_i,\,i=1..3$. 
    \textbf{(b)}: The all similar triangles found in the local trunk map, 
    \textbf{(c)}: the corresponding triangles found in Global trunk map }
    \label{fig:feature_topo_descriptor_matcher}
\end{figure}

\subsection{Matching Local and Global DT}
\label{sec:match_topo}

The DT graph obtained from the global point cloud map, $G_{map} = (V, E, T)$, defines a set $V$ of vertices, a set $E\subseteq V^2$ of edges and a set $T\subseteq V^3$. Each vertex represents a trunk in the point cloud map. Edges establish a relation between a point and its natural neighbors it is connected within the graph. For each local DT graph that we obtain by aggregating consecutive lidar frames in real-time, our objective is to find a matching subgraph for $G_{local}$ in the graph $G_{map}$. We proceed as follows in order to obtain such subgraph.

\textbf{Triangle search based on dissimilarity.} A dissimilarity $d(.,.)$ of two triangles $t_1$ and $t_2$ has two properties: it is always positive, $0\le d(t_1,t_2)$, and zero in case of identity, $d(t_1,t_1)\equiv 0$. Typical triangle dissimilarities use an intermediate vector of two descriptors, and some sort of norm between these vectors. For instance, in~\cite{arzoumanian2005} the authors utilize the ratio of the lengths of the shortest and longest edges, and the angle between those edges. We utilize the intermediate vector $(A,l)$ of a triangle $t$ with an area $A$ and $l=L^2$, where $L$ is the perimeter length. The vector components have the dimensionality of area (in square meters). Then, the dissimilarity $d(.,.)$ is defined as: $d(t_1,t_2)= |A_2-A_1|+|l_2-l_1|$.

To speed up the real-time matching process, all triangle perimeters and areas of the global DT graph have been computed offline. In terms of comparing triangles based on the magnitude of the difference of $(A,l)$ vectors, the triangles composed by peripheral points in a graph are usually different because of fewer observed points, so only triangles $T_{selected}$ which not include a peripheral point in the local graph $G_{local}$ are selected. Therefore, a set of candidate triangles $\{T_{selected} \}$ are selected and used to find a matching subset in $G_{map}$. From $\{T_{selected} \}$, we build the corresponding graph

There may exist multiple closely similar triangles $T_{candidate}$ in $G_{map}$ to any specific selected triangle $T\in\{T_{selected} \}$ in $G_{local}$. Thus the next step is to compare the neighboring triangles too. As we have excluded the peripheral triangles, the global subset $G_{selected}$ that will be the match of $\{T_{selected} \}$ should also have three neighbor triangles. The triangle neighborhood vote is performed using the same dissimilarity measure. As we show in Figure~\ref{fig:feature_topo_descriptor_matcher}, the triangle $S_0$ is one of the selected triangles 
$S_0\in T_{selected}$ in the local graph $G_{local}$, and the three triangles $S_1,S_2,S_3$ are its neighbors. 

A feature vector of the triangle $S_0$ used in the match process is combined from the intermediary dissimilarity vectors of the triangle $S_0$ and its neighbors (called a triangle star) according to \eqref{eq:triangleFeatures}.
\begin{equation}
    features(S_0) = [A_0,l_0,A_1,l_1,A_2,l_2,A_3,l_3]
    \label{eq:triangleFeatures}
\end{equation}

By comparing the $features(S_0)$ of a triangle $S_0$ to feature vectors in $T_{selected}$, a set of matching triangles $T_{candidate}$ meeting the error tolerance may be found. After a pair of similar triangles $(S_0,S_0')\in T_{selected}\times T_{candidate}\subset G_{local}\times G_{map}$ are found, the next would be to estimate the position $p_t$ and the orientation $\theta$, which  matches $T_{selected}$ with $G_{map}$.

\textbf{Calculation of corresponding vertex pairs.} In order to describe the matching process, we use the following notation. A triangle $S_0= ABC$ consists of vertices $A$, $B$, and $C$, and an edge vector $\vec{e}=\vec{AB}$ of an edge $e=AB$ is oriented and signified with its end points. 

\begin{figure*}[t]
    \centering
    \includegraphics[width=0.8\textwidth]{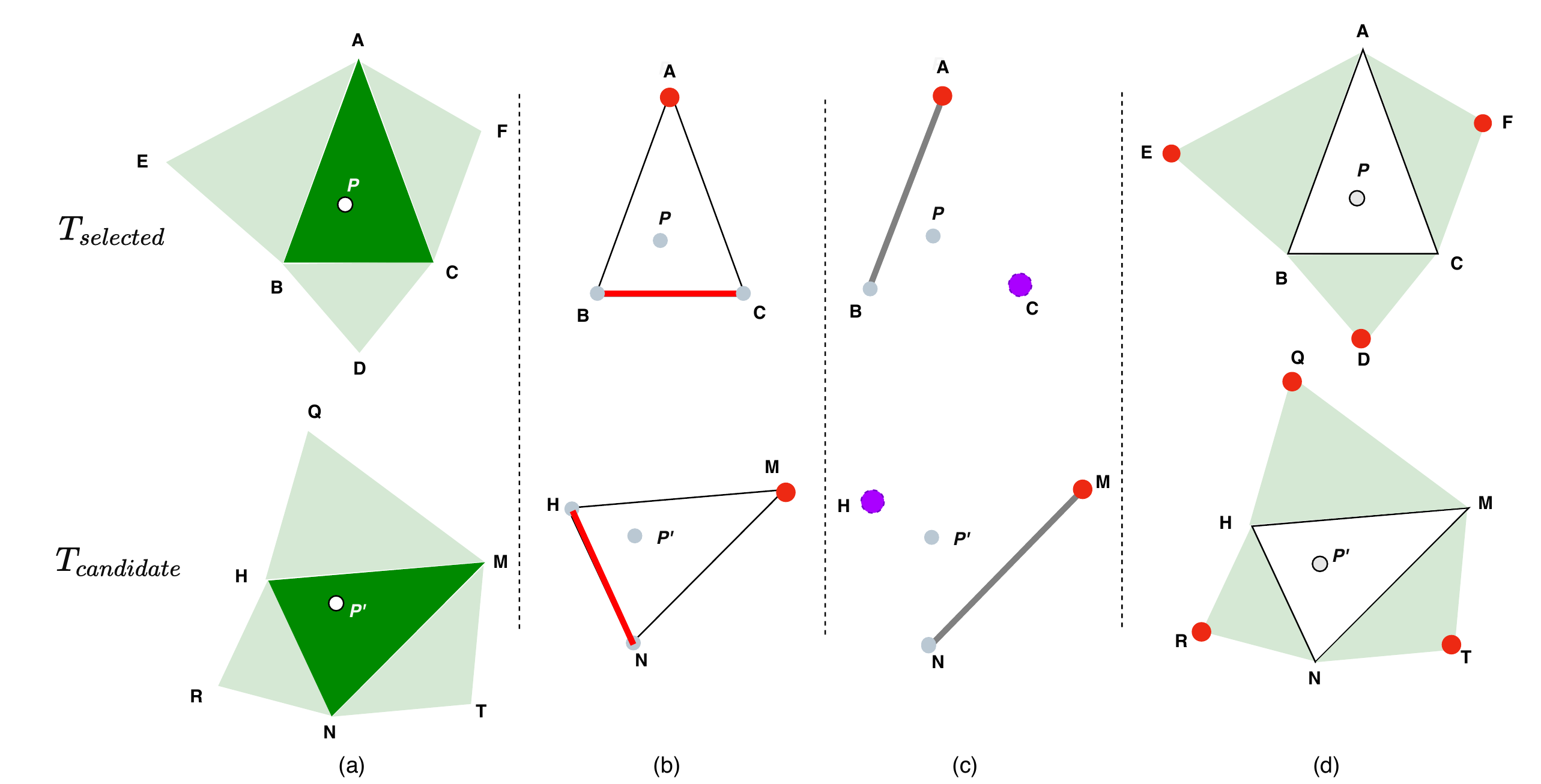}
    \caption{\textbf{(a)}: The selected triangle $T_{selected}$ in $G_{local} $ with candidate matched triangle $T_{candidate}$ in $G_{map}$
    \textbf{(b)}: Find the first correspond point $ A $ in $T_{selected}$ with it's $ M $ in $T_{candidate}$ by finding the similar featured side length
    \textbf{(c)}: Find the rest points $B$, $c$ and it's correspond points $H$, $N$.
    \textbf{(d)}: Find the correspond neighbor points $E$, $D$ , $F$ and it's correspond points $Q$, $R$, $T$.}
    \label{fig:correspond_match_points_process}
\end{figure*}

To solve the transformation parameters $p_t$ and $\theta$, we first need to find the correspondence between vertices of triangle stars $\mathcal{S} = \{S_i\}_{i=0..3}\in T_{selected}$ and triangles $\mathcal{S}'= \{S_i'\}_{i=0..3}\in T_{candidate}$. The definition of a triangle star is easily seen from Figure~\ref{fig:correspond_match_points_process} (a). As the figure shows, the vertex match between $T_{selected}$ and $T_{candidate}$ can be divided into three steps. The first step is in the detail (b). We then find the first matching pair $S_0= ABC\in T_{selected}$ and $S_0'= MHN\in T_{candidate}$. Through comparing the side lengths between two triangles, an edge $BC$ of the $ABC$ can be selected
so that there is an edge $|NH|$ with similar length in the $MHN$, so the remaining vertices $A$ and $M$ are a pair of 
corresponding vertices. For the second step, the goal is to find the other corresponding vertices in $ABC$ and $MHN$. 
As Figure~\ref{fig:correspond_match_points_process} $(c)$ shows, selecting one edge which has the vertex $A$ as already known, and one edge contains the vertex $M$ in the $\triangle MHN$, then we compute which side of the edge the remaining vertices $C$ and $H$ are located. This happens by inspecting the value $a\in\mathbb{R}$ of the following formula:

\begin{equation}
    a= (\Vec{CA} \times \Vec{CB}) \cdot (\Vec{HM} \times \Vec{HN})
    \label{eq:V}
\end{equation}

If $a > 0$, the vertices $C$ and $H$ are on the same side of the edge $AB$ and edge $MN$, otherwise the two vertices are on the opposite. In the case shown in Figure~\ref{fig:correspond_match_points_process} $(c)$, the result is $a < 0$, so the vertex pairs are $(C,M)$ and $(B,H)$. The last step is to match vertices in the triangle neighbors. Since we already know the correspondence between triangles $ABC$ and $MHN$, we just need to get which two vertices are included in the neighboring triangles. For example, as we get the corresponding pairs of vertices $(A,M)$ and $(B,H)$ from last two steps, the last vertices $E$ and $Q$ are a match, too. 

The match between vertices of two triangle sets is now complete. The match relation is one-to-one, so it can be recorded 
as a permutation function: $f:\,[1,6]\subset\mathbb{N}\rightarrow [1,6]\subset\mathbb{N}$. This way each vertex $V_i\in T_{selected}$ has a matching vertex $V_{f(i)}\in T_{candidate}$. Next we will use the permutation $f$ while estimating the best possible translation and rotation.

\subsubsection{Rotation and Translation Estimation}

By comparing the vertex orientations in the found two triangle sets $\mathcal{S}=\{S_i\}_{i=0..3}\in T_{selected}$ and  $\mathcal{S}'=\{S_i'\}_{i=0..3} \in T_{candidate}$, we can find a unifying 2D rigid body transformation $T[p_t, \theta]$ between the local sample and the global map. The transformation $T$ consists of a rotation of the angle $\theta$ followed up with a translation $p_t$.

To calculate the transformation parameter $p_t$ we utilize the definitions in \eqref{eq:solveTArgs}, where the translation $t_p$ is the best possible one estimate, since it equates the mean of two patterns. The angle $\theta$ is defined by measuring how large a rotation is needed to transform a vectors $V_i - \overline{V}$ to $V_{f(i)}'-\overline{V}'$, where the $\overline{V}$ represents the mean of $T_{selected}$ and $\overline{V'}$ is the mean of the $T_{candidate}$.

There are six pairs $i=1...6$ of vertices coupling $T_{selected}$, $T_{local}$ and $T_{candidate}$ from $G_{map}$, and six possible candidates $\beta_i$ to be chosen as the final local rotation $\theta$. We choose the value of $\beta_i$ which fits the triangle patterns $T_{selected}$ and $T_{local}$ with the least error. One important detail is arranging a signed angular measure between two vectors. We use a rectangular rotation matrix $P=\big(\begin{smallmatrix} 0 & -1 \\ 1 & 0 \end{smallmatrix}\big)$ to get a positive sign $sign(Pa\cdot b)$ for rotation angles, which move a vector $a$ to a vector $b$. We omit possible fit minimizing values in between the angles $\{\beta_i\}_{i=1..6}$ because the objective at this stage is to only have a close estimate and not the final position estimation. 

Using a zeroth power as a shorthand when producing unit vectors $a^0= a/\|a\|$, we define counter-clockwise oriented angles $\beta_i$, which rotate vectors $V_i$ parallel to vectors $V_{f(i)}$. Finally, we define approximate values for the translation $p_t$ and the rotation $\theta$ according to \eqref{eq:solveTArgs}.

\begin{equation} 
    \begin{cases}
        & \overline{V}= {~\mathbf{mean}~}_{V\in T_{selected}} V \\[+2ex]
        & \overline{V}'= {~\mathbf{mean}~}_{V'\in T_{candidate}} V'  \\[+2ex]
        &  p_t =  \overline{V}' - \overline{V}   \\[+2ex]
        & \beta_i= sign\left(PV_i\cdot V_{f(i)}\right)\arccos{\left( V_i^0\cdot V_{f(i)}^0 \right)},\:\:i=1..6 \\[+2ex]
        &  
        \begin{split} 
        & \theta = {~\mathbf{argmin}~}_{\beta_i,\:i=1\dots6} \\
        & \displaystyle\sum_{j=1\dots6} \left\Vert
        \begin{bmatrix}
            \cos{\beta_i} & -\sin{\beta_i} \\
            \sin{\beta_i} & \cos{\beta_i}
        \end{bmatrix}
         \left[ V_j - \overline{V} \right]^T  - 
         \left[ V_{f(j)} - \overline{V'} \right]^T 
        \right\Vert\\ 
        \end{split}
    \end{cases}\label{eq:solveTArgs}
\end{equation} 
 
\subsubsection{Geometric Verification, Final Translation and Rotation Estimation}

From the $G_{local}$ match against $G_{map}$, usually multiple matches $\{(\mathcal{S}_k,\mathcal{S}_k')\}_{k=1..m}$ between  $T_{selected}$ and $T_{candidate}$ can be found. Therefore, we need to select one among $m$ transformation candidates with rotation and translation parameters $\theta_k$ and $t_k$. This final step is to find the best estimation between $G_{local}$ and $G_{map}$. Let the corresponding points of a match candidate be $(V_{local}, V_{map})$. Starting from the previously computed initial guesses $\theta$ and $p_t$, we select the final solution from \eqref{eq:finalMatch}. Note that this does not ensure global convergence, and unexpected results might be obtained if the global map is too homogeneous, not enough trees are detected, or the map is large and multiple positions yield similar error. Through our experiments, nonetheless, we have been able to confirm stable position estimation when enough consecutive lidar scans were aggregated.

\begin{equation}
\begin{split}
    & (\theta, t_x,t_y) = \\ 
    & \arg_{\beta, x_t,y_t}\min_{\substack{\beta\in B\\x\in D_x\\y\in D_y}} 
   \sum_{j=1..m} 
    \left\Vert 
     \begin{bmatrix}
        \cos{\beta} & -\sin{\beta} & x\\
        \sin{\beta} & \cos{\beta} & y \\
        0 & 0 & 1
    \end{bmatrix}
    \left[ V_j^T - V_{f(j)}^T \right]
    \right\Vert \label{eq:finalMatch} 
\end{split}
\end{equation}

where $m$ is the number of local matches, $\beta \in B = [\min{\theta_j}_{j=1..m}, \max{\theta_j}_{j=1..m}] \subset \mathbb{R}$ goes through the scope occurring in the local rotation angles, and $x$ goes through a similar scope of $D_x$ occurring in translations along the x axis. $D_y$ is defined correspondingly on the $y$ axis. Note that both \eqref{eq:solveTArgs} and \eqref{eq:finalMatch} require the vertices to be in a homogeneous form $V\rightarrow [V\; 1]$. The local start point is $(x_0,y_0)$ with an orientation of $\theta_0$, and the final estimation of the robot position is $(x_0 + t_x, y_0 + t_y)$ with an orientation $\theta$ in the global map.

\section{Results}

This section presents our experimental results. We analyze the performance of our method from the point of view of the trunk segmentation as well as the DT matching between the local graph and a subset of the global graph.

\begin{figure*}
    \centering
    \includegraphics[width=0.8\textwidth]{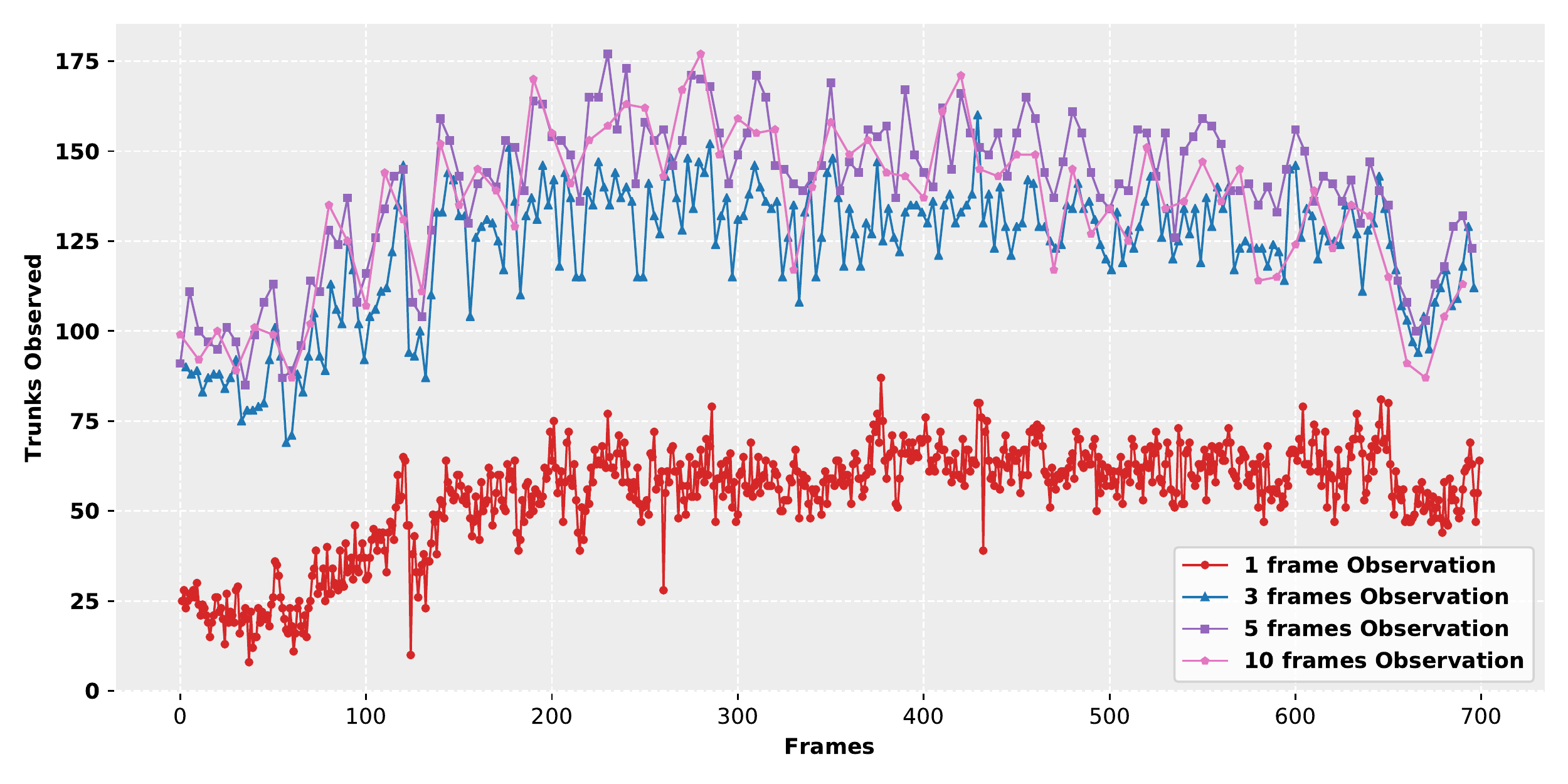} 
    \caption{Number of trunks observed for different number of consecutive aggregated frames.}
    \label{fig:trunk_observed_plot}
\end{figure*} 

\begin{figure*}
    \centering
    \includegraphics[width=0.8\textwidth]{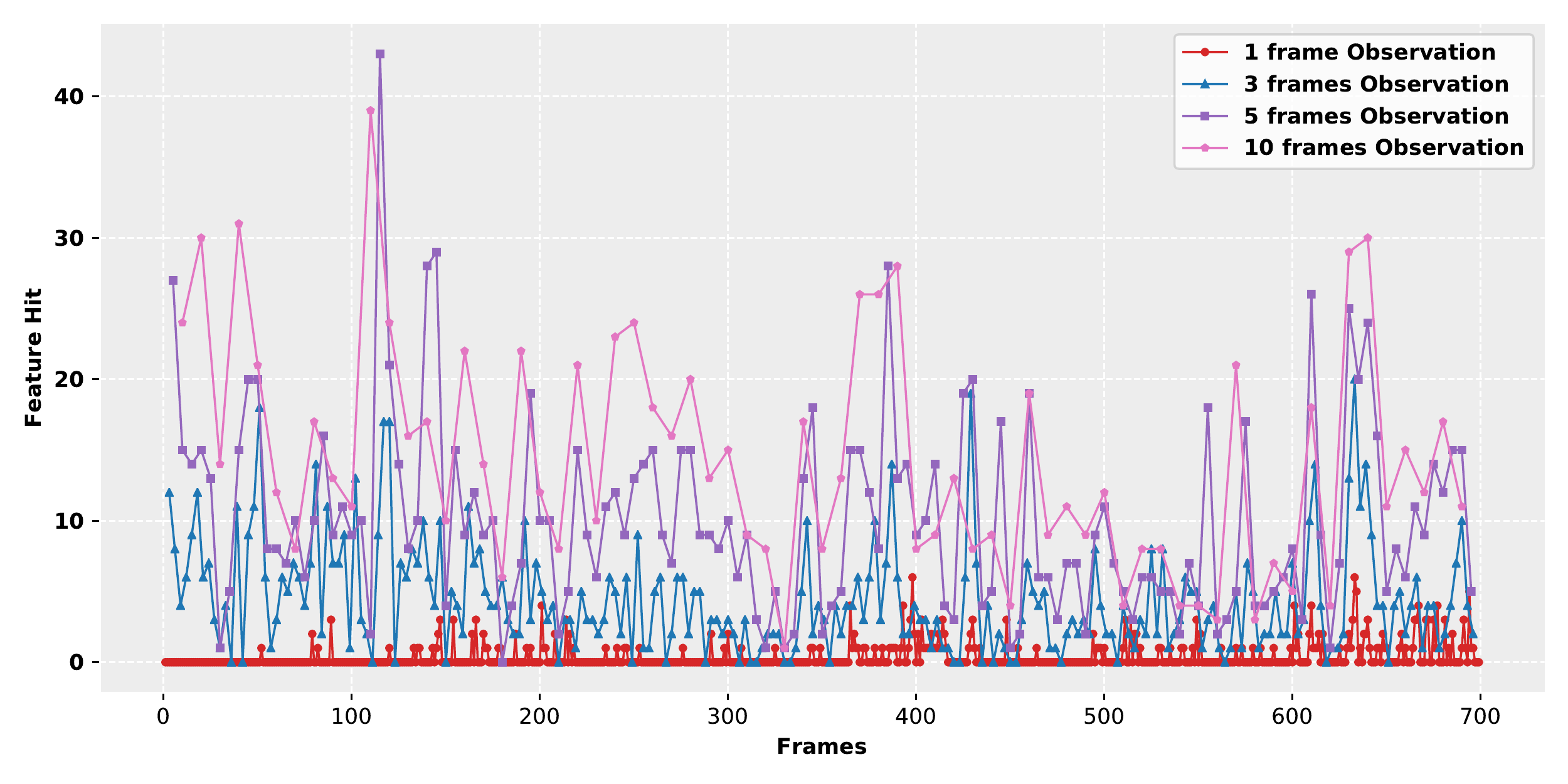} 
    \caption{Number of trunks matched between $G_{local}$ and $G_{map}$ for different number of consecutive aggregated frames.}
    \label{fig:feature_hit_plot}
\end{figure*} 

\begin{table*}
    \caption{Number of trunks detected and success match for different number of consecutive aggregated frames (average number of trunks matched per scan and total ratio of successful matches).}
    \centering
    \begin{tabular}{@{}cccc@{}}
        \toprule
        \centering\textbf{\#Aggregated Frames} & \centering\textbf{Average \#Trunks} & \centering\textbf{Average \#Matching Triangles}	& \textbf{Overall Match Success Rate} \\
        \midrule
        \centering 1		& \centering 54.09		    & \centering 1.71          & 22.89\,\%   \\
        \centering 3		& \centering 122.98  		& \centering 4.97          & 88.79\,\%   \\
        \centering 5 		& \centering 139.54         & \centering 10.22         & 99.28\,\%   \\
        \centering 10	    & \centering 133.96         & \centering 14.37         & 100\,\%     \\
        \bottomrule
    \end{tabular}
    \label{tbl:trunk_detected_table}
\end{table*}

\subsection{Trunk Segmentation Performance}

Figure~\ref{fig:trunk_observed_plot} shows the number of trunks that can be observed from a single lidar scan, and when aggregating three, five or ten consecutive frames. In Table~\ref{tbl:trunk_detected_table} we show the average number of trunks observed, as well as the average number of trunks that can be matched with the global map after generating the DT graph. We also include the overall matching success rate, which is defined as the number of local DT graphs that have been successfully matched with the corresponding subsets of the global DT graph. From individual lidar frames, we were only able to obtain an average of 54.09 valid trunks in total. When aggregating consecutive scans, we were able to obtain approximately a twofold increase in the number of detected trunks, and a tenfold increase in the number of matched trunks. The number of aggregated frames thus has a significant impact on the performance of our algorithm. As the number of trunks and their positions determines the appearance of the local DT graph, this will directly affect the number of features detected by our descriptor. The number of trunks observed is the key factor influencing the overall results of our algorithm. From Table~\ref{tbl:trunk_detected_table} we can see the relationship between the DT matching success and the number of trunks. With a single frame observation, only an average 1.71 trunks are successfully matched; with three aggregated frames, there are in average 4.97 trunks matched; with 5 and 10 aggregated frames, there are more than 10 trunks in average that our algorithm can match with the global map, taking the overall DT match success rate to 100\,\% in the latter case.

\subsection{Computational Time and Accuracy}

\begin{figure}[t]
    \centering
    \includegraphics[width=0.5\textwidth]{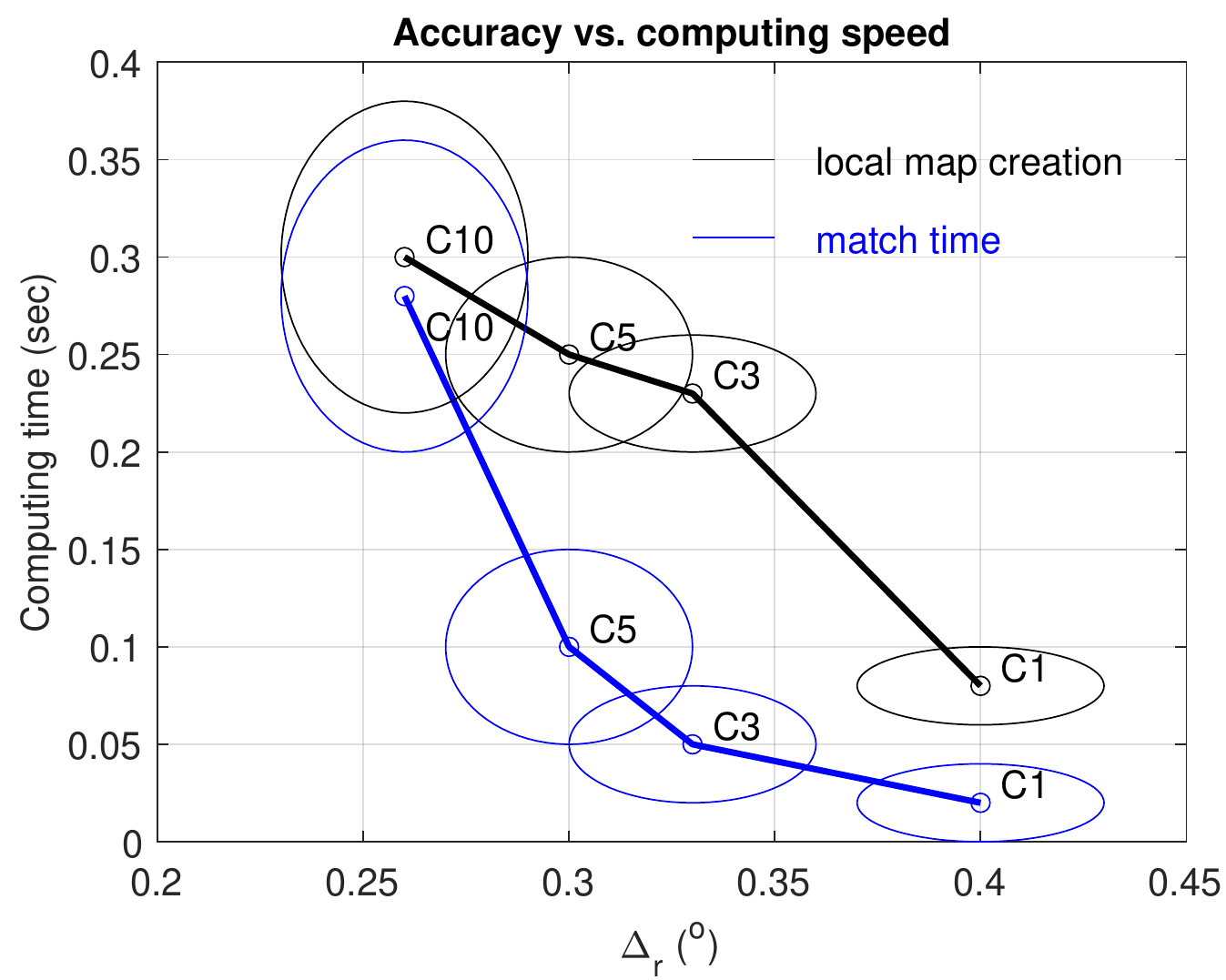}
    \caption{Dependency of the rotational accuracy
    and computation time on the choice of the number
    of aggregated frames. Both the local map
    creation time (black) and the similarity
    matching time (blue) are depicted. Number of 
    frames ranges from 1 to 10 (C1, C3, C5, C10).
    Ellipses represent the s.t.d.'s of the measurements.}
    \label{fig:time_accuracy}
\end{figure}

The key matching process is based on the calculation and matching of the Delaunay triangulations, which is in turn based on the aggregation of 1, 3, 5 or 10 lidar frames. These four possibilities with a varying number of aggregated frames are labeled as C1, C3, C5 and C10 in Figure~\ref{fig:time_accuracy}. The aggregation of frames adds complexity and requires computing time both in the local map creation phase (black line) and the similarity matching time (blue line). We can see that the rotational location error significantly improves when aggregating more frames. The computing time of the match process grows approximately in terms of $O(\Delta_r^{-2})$, where $\Delta_r$ is the angular accuracy achieved by different choices C1, C3, C5, C10. In summary, the computation time is about 0.1\,seconds for a single frame (C1) and 0.6\,seconds when aggregating 10 frames (C10).

To test our pose estimation accuracy, we used the original LOAM position as the ground truth. Figure~\ref{fig:time_accuracy} depicts only the rotational error, since the translation error was approx. 0.2\,m with an standard deviation of 0.14\,m in all cases. It is worth noting that the rotational error is not cumulative, since the match between the local and global DT graphs is being done separately and independently at each location. The previous location is only used as the initial state. The maximum occurred translation error was approx. 0.5\,m and the maximum rotation error \ang{2.23}, both for the C1 scenario.  

We can also see that the match time is relatively stable in the C1-C5 scenarios. The local map creation time (aggregation of frames) stabilizes as the number of frames increases, owing to the stabilization in the number of trees that can be observed. The segmentation algorithm is implemented in C++, while the matching process is implemented in Python. 
\section{Discussion}
An application topic of the proposed navigation system is an automated transport robot, which moves logs from the forest to the collection road. Since work happens in two phases, where a forest harvester produces stripping roads and fells trees, and the collection phase, a robot can assume to have the global map built by the harvester which is not necessarily autonomous. This way its task is just to orientate itself along the strip road. A challenge arises from the high utility load of the transfer robot, which can be up to 8000\,kg. A transfer robot may not have an as big capacity, but the ability to locate itself, and to detect possible stability risks is a requirement. 

Also, a transfer robot moves faster than the harvester going through its work cycle. A good estimate in rough terrain is 1.0\,m/s. Thus, a fast and accurate odometric computation is essential.  

\subsection{Topology mapping}

In this paper, the forest trunk topology map is generated from the previous lidar data by the harvester. There are also two alternative ways to generate the global map: a large-scale (e.g. nationwide) digital forest inventory, or an aerial laser scan (ALS) from a drone working as an autonomous team member of the harvester and the robot carrier. Compared to the point cloud map, the topology map has the potential to do fast localization and dynamic update without much computation burden and makes a 3 actor teams (harvester-drone-robot) an interesting target of further study. 

\subsection{Potential accuracy improvements} 

From the experiments, we can see that the number of trunks detected in each observation significantly influence the result of the success matching ratio. In our method, due to the LIDAR sensor characteristic, far trunks are unable to be recognized because of too few LIDAR points sampled from the appearance. Therefore, we can employ other sensors like the depth camera or RGB camera which can obtain more detail of the trunk appearance, which lets the system get more capability to recognize the trunks from PC. 

The other significant error exists in the trunk position estimation. In our method, the $G_{map}$ and $G_{local}$ are generated from the segmented trunks points in $PC_{map}$ and $PC_{local}$ by calculating the average coordinate of points. However, it's impossible to get each individual trunk whole appearance observation with several consecutive lidar observations. As the observation in different positions will get different point cloud data of different sides of the trunk, the estimation pose of the trunk will be different in each local graph. However, the global graph has a more accurate trunks position as more details of each trunk collect from different sensors and different views, so the more accurate trunk position estimation in the local map can increase the chance of successful matching between local graphs and global graph. To reach a more robust system, we also can utilize other sensors like a camera through sensor fusion to get more details about each individual trunk.  

\subsection{Potential computational improvements}

We proposed an efficient, robust framework to locate the robot harvester in large area forest and we tested it on a real harvester. In our case, the GNSS/GPS info is not taken into account. The location accuracy is approx. 0.3\,m from consequential field measurements and approx. 2\,m, when comparing to general odometric result. This accuracy is enough to give a robot an initial position, and this can accelerate the matching process by conveniently initializing the search with a close match. For a real application, the GPS also can help decide which global point cloud map it will access from the cloud server for the localization.

The local Delaunay triangularization generating the local trunk map $T_{local}$ can be implemented in a radius-limited way by using the S-Hull method~\cite{sinclair2016}, which limits the size $n$ of the computational task with the well-established complexity $O(n\log{n})$. A selection of an active subset of triangles can be made in the global map $T_{local}$ based on the previous localization result. Together these improvements have the potential to make the matching process much faster. 

There is a possibility to speed the convergence of the search of the final transformation in \eqref{eq:finalMatch} by using a branch-and-bound algorithm~\cite{boukouvala2015} with properly set estimates for local extreme. This possibility is left for future research. 

All in all, the C1-C5 cases are applicable to the intended situation where the robot moves at approx. 1\,m/s and has a sampling rate 2\,scans/sec. The case C10 is within the reach after implementing the above-mentioned improvements.
\section{Conclusion}

In this study, we proposed a simple yet effective segmentation-based approach to detect trunk position and Delaunay Triangulation(DT) geometry-based localization method for autonomous robots navigating in a forest environment. The proposed methods can provide accurate positioning based only on real-time lidar data processing in the unstructured and relatively complex environments that forests represent. The proposed method can be utilized for harvesters or other autonomous robots enabling fast global localization and recognizing individual trunks in real-time.   

Our experiments show the proposed method reach accurate global localization precision without a good initial pose or GPS signal. The proposed method is simple and efficient, and it is a sensible solution to meet localization needs of harvesting operations in the forest. In future work, we plan to explore the forest localization algorithm in the context of significantly larger forests and to apply the proposed method at a system level for map updating or within the SLAM stack.

\section*{Acknowledgment}
he data from the test site was gathered in 
co-operation with Stora Enso oy, Mets\"{a}teho oy, and Aalto University.This research are funded by Business Finland grant number 26004155 and Academy of Finland grant number 328755

\ifCLASSOPTIONcaptionsoff
  \newpage
\fi



%
\bibliographystyle{unsrt}
\bibliography{main.bib}

%








\end{document}